\documentclass[letterpaper]{article} 
\usepackage{aaai2026}  
\usepackage{times}  
\usepackage{helvet}  
\usepackage{courier}  
\usepackage[hyphens]{url}  
\usepackage{graphicx} 
\usepackage{natbib}  
\usepackage{caption} 
\usepackage{xcolor} 
\usepackage{amsfonts}
\usepackage{bm} 
\usepackage{dsfont} 
\usepackage{subcaption} 
\usepackage{algorithm, algpseudocode} 
\usepackage{amsmath, amssymb}
\usepackage{booktabs} 
\usepackage{siunitx}  
\usepackage{pifont}
\usepackage{xspace}

\algnewcommand\algorithmicinput{\textbf{Input:}}
\algnewcommand\algorithmicoutput{\textbf{Output:}}
\algnewcommand\Input{\item[\algorithmicinput]}
\algnewcommand\Output{\item[\algorithmicoutput]}

\newcommand{\ModelName}{Zo3T\xspace}

\usepackage{newfloat}
\usepackage{listings}
\DeclareCaptionStyle{ruled}{labelfont=normalfont,labelsep=colon,strut=off} 
\floatstyle{ruled}
\newfloat{listing}{tb}{lst}{}
\floatname{listing}{Listing}
\title{\ModelName: Zero-Shot 3D-Aware Trajectory-Guided Image-to-Video Generation \\ via Test-Time Training}
\author {
    Ruicheng Zhang\textsuperscript{\rm 1}\footnote{Equal contribution.},
    Jun Zhou\textsuperscript{\rm 1}\footnotemark[1] \footnote{Corresponding author.},
    Zunnan Xu\textsuperscript{\rm 1}\footnotemark[1],
    Zihao Liu\textsuperscript{\rm 1},
    Jiehui Huang\textsuperscript{\rm 2},\\
    Mingyang Zhang\textsuperscript{\rm 3},
    Yu Sun\textsuperscript{\rm 4},
    Xiu Li\textsuperscript{\rm 1}\footnotemark[2]
}
\affiliations {
     \textsuperscript{\rm 1} Tsinghua University\\
     \textsuperscript{\rm 2} Hong Kong University of Science and Technology\\
    \textsuperscript{\rm 3} China University of Geosciences\\
    \textsuperscript{\rm 4} Sun Yat-sen University
}

\usepackage{bibentry}

\begin{document}

\maketitle

\begin{abstract}
Trajectory-Guided image-to-video (I2V) generation aims to synthesize videos that adhere to user-specified motion instructions. Existing methods typically rely on computationally expensive fine-tuning on scarce annotated datasets. Although some zero-shot methods attempt to trajectory control in the latent space, they may yield unrealistic motion by neglecting 3D perspective and creating a misalignment between the manipulated latents and the network's noise predictions. To address these challenges, we introduce Zo3T, a novel zero-shot test-time-training framework for trajectory-guided generation with three core innovations: First, we incorporate a 3D-Aware Kinematic Projection, leveraging inferring scene depth to derive perspective-correct affine transformations for target regions. Second, we introduce Trajectory-Guided Test-Time LoRA, a mechanism that dynamically injects and optimizes ephemeral LoRA adapters into the denoising network alongside the latent state. Driven by a regional feature consistency loss, this co-adaptation effectively enforces motion constraints while allowing the pre-trained model to locally adapt its internal representations to the manipulated latent, thereby ensuring generative fidelity and on-manifold adherence. Finally, we develop Guidance Field Rectification, which refines the denoising evolutionary path by optimizing the conditional guidance field through a one-step lookahead strategy, ensuring efficient generative  progression towards the target trajectory.
Zo3T significantly enhances 3D realism and motion accuracy in trajectory-controlled I2V generation, demonstrating superior performance over existing training-based and zero-shot approaches. 
\end{abstract}

\begin{links}
    \link{Home Page}{https://richard-zhang-ai.github.io/}
\end{links}

\section{Introduction}
Recent advances in text- and image-driven video diffusion models have demonstrated remarkable capabilities in generating photorealistic and semantically coherent videos~\cite{ho2022imagen,blattmann2023stable,zhou2025fireedit,yuan2025identity,chen2023videocrafter1,xu2025hunyuanportrait,ma2025followfaster}. These developments have paved the way for controllable object motion animation~\cite{he2024cameractrl,namekata2024sg,levitor,ma2025followyourmotion}, which plays a vital role in practical applications such as virtual reality, gaming, advertising, and digital art~\cite{ma2025controllable}.

\begin{figure}[t]
\centering
\setlength{\abovecaptionskip}{-0.03em}   
\includegraphics[width=1\columnwidth]{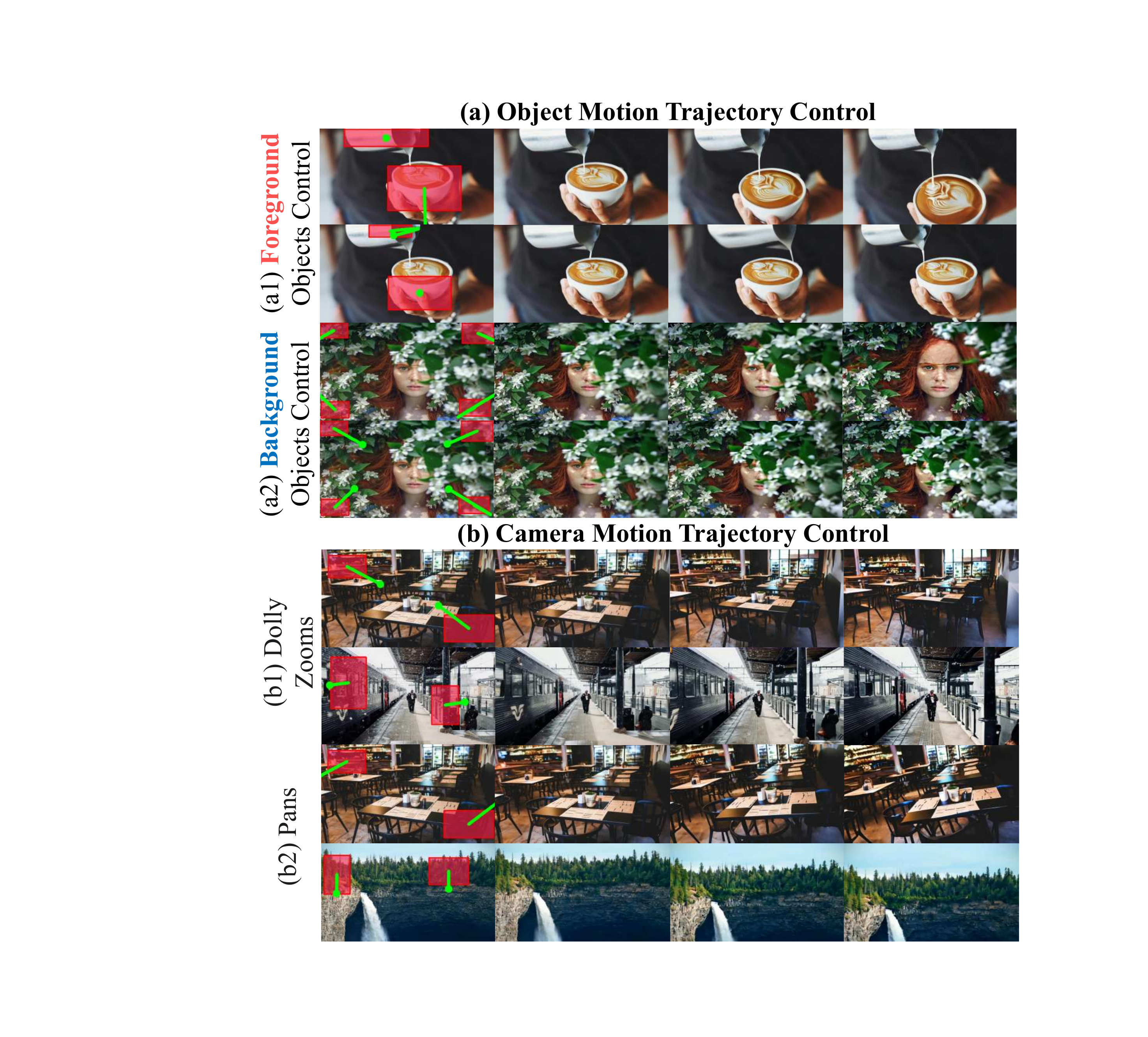} 
\caption{Versatile Trajectory Control with Our Method. Given a set of bounding boxes with corresponding trajectories, our framework enables precise control over diverse object and camera motions. By leveraging the inherent knowledge of a pre-trained video diffusion model, we achieve zero-shot trajectory guidance without any fine-tuning.}
\label{fig:teaser}
\vspace{-2em}
\end{figure}

\begin{figure*}[t]
\centering
\setlength{\abovecaptionskip}{-0.03em}   
\includegraphics[width=1.65\columnwidth]{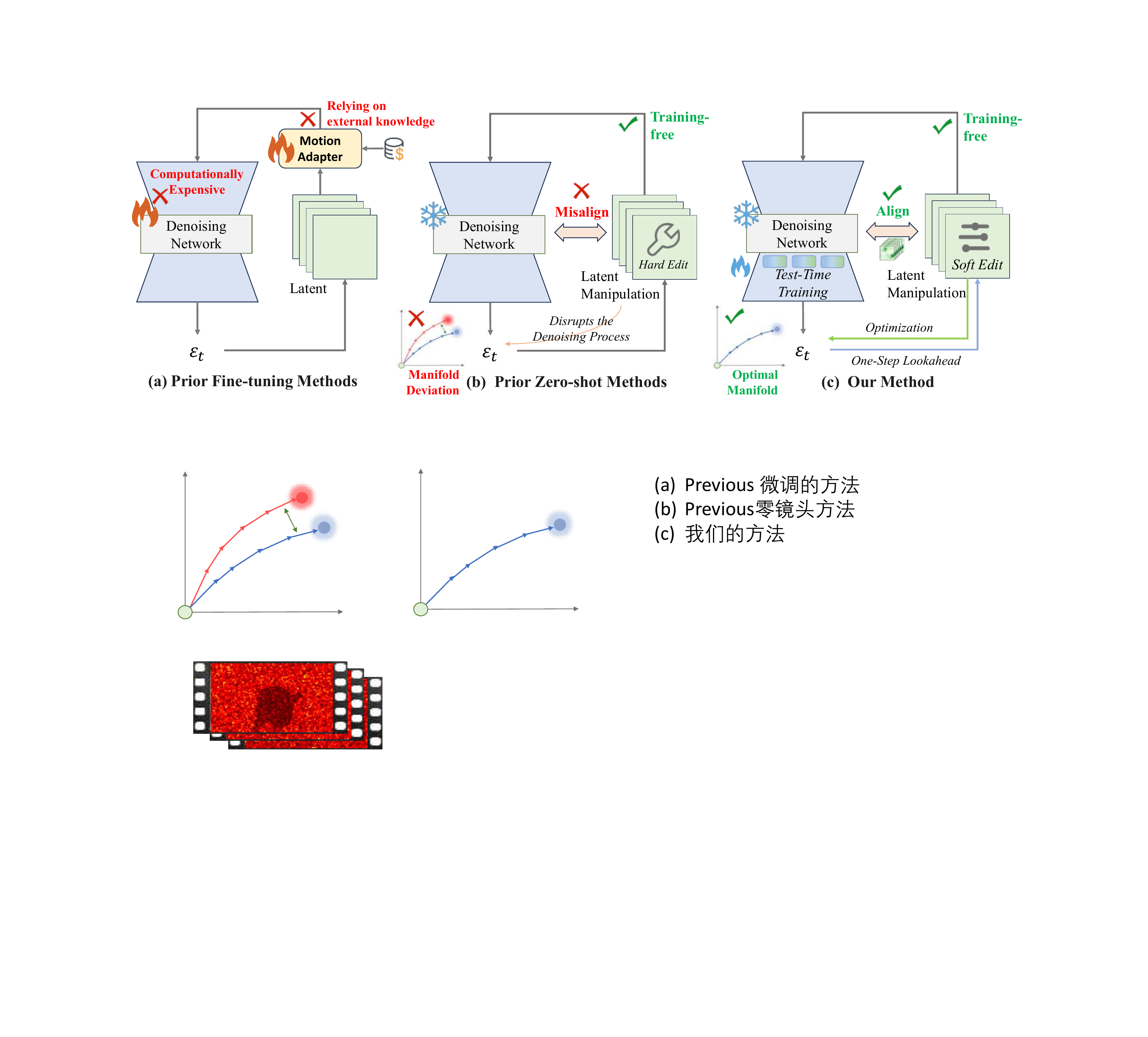} 
\caption{Advantages of our Method over prior works.}
\label{fig:model_compare}
\vspace{-1.8em}
\end{figure*}

Early works~\cite{wu2024draganything, wang2024motionctrl,kong2024hunyuanvideo} motion-controllable video generation employed annotated bounding boxes or trajectory points to fine-tune video diffusion models. However, their high computational cost severely limits their practical applicability. In contrast, training-free approaches have  attracted significant attention due to their efficiency. These methods modify either the latent representations of target objects through attention mechanisms~\cite{ma2023trailblazer, namekata2024sg} or the noise construction~\cite{qiu2024freetraj} during the denoising process, thereby enabling accurate motion control and visually coherent results. However, this paradigm of direct latent space manipulation harbors a fundamental tension. The pre-trained denoising network $\epsilon_\theta$ has learned a delicate mapping from a specific distribution of noisy latents to their corresponding noise predictions. Aggressively editing $\mathbf{z}_t$ to enforce motion pushes it ``off-manifold", creating a misalignment between the manipulated latent and the model's learned prior. This \textbf{model-data misalignment} forces the network to denoise a state it is never trained to see, frequently causing a catastrophic degradation in visual quality, manifesting as textural collapse, loss of identity, and other temporal artifacts~\cite{he2023manifold, ReNoiseInversion}. 
Moreover, most existing approaches guide motion using 2D trajectories paired with fixed-size bounding boxes or masks. This representation suffers from inherent ambiguity, as it fails to capture the perspective scaling an object should undergo as its depth changes. As shown in Figure~\ref{fig:ablation_qualitative}(a), this lack of 3D awareness leads to physical implausibility and visual distortions, such as unrealistic scaling and motion patterns.

To address these limitations, we propose 
\textbf{\ModelName}, a novel Test-Time Training (TTT) framework for zero-shot image-to-video (I2V) generation with 3D-aware trajectory control. To mitigate the model-data misalignment in the trajectory guidance process, we introduce a ``soft-editing" strategy that co-adapts the data and the model through Trajectory-Guided Test-Time Training (Figure~\ref{fig:model_compare}). This mechanism injects and optimizes an ephemeral LoRA module within the denoising network at inference time. 
A trajectory-consistency loss simultaneously updates both the latent state $\mathbf{z}_t$ and the temporary LoRA weights. This co-adaptation realigns the model to the edited latent, ensuring the denoising process remains on a stable, high-fidelity manifold while adhering to the specified motion. Building on this, we further introduce Guidance Field Rectification (GFR) via re-evaluating noise scores, which adjusts the guidance field adaptively by maximizing the generation of instance features in the trajectory motion area at each denoising step to adaptively adjust the guidance field. This optimization encourages the latent features to evolve toward the desired trajectory, ensuring controllable and high-quality video synthesis.
To ensure physical plausibility, \ModelName projects user-specified 2D trajectories into a 3D space, leveraging depth cues. This enables perspective-correct motion, ensuring realistic scaling and movement. While recent I2V methods like LeviTor~\cite{levitor} and ObjCtrl-2.5D~\cite{objctrl2.5d} also incorporate 3D information, LeviTor requires costly training on extensive masked datasets, and ObjCtrl-2.5D's core assumption of equating object motion with camera movement constrains its applicability in complex scenes. In contrast, our approach offers greater flexibility and directness, requiring no external motion priors or restrictive assumptions.

By jointly leveraging our 3D-aware trajectory control, test-time adaptation strategy, and guided field rectification module, ours framework achieves precise motion control and camera movement, as shown in Figure~\ref{fig:teaser}. 
Extensive experiments validate the effectiveness of \ModelName in generating visually coherent video sequences across a diverse range of trajectory control scenarios in a zero-shot setting.
In summary, the main contributions of our framework can be summarized as follows:
\begin{itemize}
   \item We propose \ModelName, a 3D-aware test-time training framework for zero-shot controllable video generation that enables precise control over both target object motion and camera movement. 
   \item \ModelName integrates lightweight test-time LoRA modules during trajectory-guided generation test time to adaptively guide the generation process and maintain generative fidelity during latent manipulation. Furthermore, we refine the guidance field by re-evaluating noise scores to enforce trajectory fidelity. 
   \item Extensive experiments demonstrate that our proposed method outperforms both training-based and training-free methods in terms of trajectory control and fidelity of generated videos.
\end{itemize}

\section{Related Work}

\subsection{Controllable Generation with Diffusion Models}
The success of diffusion models~\cite{ho2020denoising, rombach2022high, song2020score} have turned focus from unconditional to fine-grained, controllable synthesis.
While extensively explored in image generation~\cite{ControlNet, GLIGEN, SpaText,zzz1}, extending such control to the video domain introduces the formidable challenge of maintaining temporal coherence while manipulating spatial content~\cite{zzz2}. The dominant approach to this problem has been supervised fine-tuning, where a large, pre-trained video model is adapted to new control modalities on specialized datasets. This paradigm includes methods conditioned on dense signals, such as per-frame pose or depth maps~\cite{zhao2023motiondirector, ma2023follow,chen2023control,jin2024alignment}, which draw inspiration from image-centric architectures like ControlNet~\cite{ControlNet}.

To offer a more intuitive user interface, a significant lineage of work has focused on sparse trajectory control~\cite{ma2023trailblazer, guo2023sparsectrl, zhang2024tora,ma2023magicstick}. Early attempts often relied on intermediate representations like optical flow~\cite{hao2018controllable}, which could introduce visual artifacts. More recent methods integrate trajectory guidance directly into the diffusion backbone. For example, DragNUWA~\cite{yin2023dragnuwa} demonstrated a framework for fusing multi-modal trajectory inputs through a multi-scale architecture. Recognizing the semantic ambiguity of a single moving point, DragAnything~\cite{wu2024draganything} advanced this concept to the entity-level by leveraging semantic features from the diffusion U-Net to guide the motion of entire objects beyond pixels. Despite their advances, these supervised methods are fundamentally constrained by the high computational cost of retraining and their dependence on large-scale, annotated video datasets, which limits their flexibility and scalability.

\subsection{Latent Manipulation for Zero-shot Optimization}

Instead of fine-tuning, an alternative approach is to manipulate the generation process at inference, leaving the model weights frozen.
This approach was first validated in the image domain~\cite{PromptToPrompt, MasaCtrl}.
Inspired by DragGAN~\cite{pan2023drag}, methods like DragDiffusion~\cite{shi2024dragdiffusion} established that precise, point-based spatial edits could be achieved by directly optimizing the noisy latent code \(z_t\) with respect to supervisory signals derived from the U-Net's internal feature maps.

However, translating this zero-shot optimization to video is non-trivial. A key obstacle is the weak temporal correspondence of feature maps within standard video diffusion models~\cite{tang2023emergent, luo2023dhf}, which makes them unreliable for guiding motion. An intuitive solution involves a modification to the self-attention mechanism to enforce cross-frame feature alignment~\cite{PromptToPrompt, MasaCtrl, geyer2023tokenflow}, thereby creating a usable guidance signal for latent optimization.
However, a more fundamental challenge pervades these test-time optimization techniques. The direct, and often aggressive, manipulation of \(z_t\) can push the latent representation off the manifold learned by the pretrained model~\cite{he2023manifold, ReNoiseInversion}. This ``manifold deviation'' disrupts the delicate denoising process, frequently causing a significant degradation in visual quality, manifesting as textural collapse, loss of identity, and other temporal artifacts. Thus, a critical open question remains: how to enforce precise, user-defined control without sacrificing the generative fidelity inherent to the foundational model. Our work directly addresses this tension by introducing a new framework that co-adapts the model and the latent state, ensuring that the guided generation remains on-manifold.

\begin{figure*}[t]
\centering
\setlength{\abovecaptionskip}{-0.03em}   
\includegraphics[width=1.\textwidth]{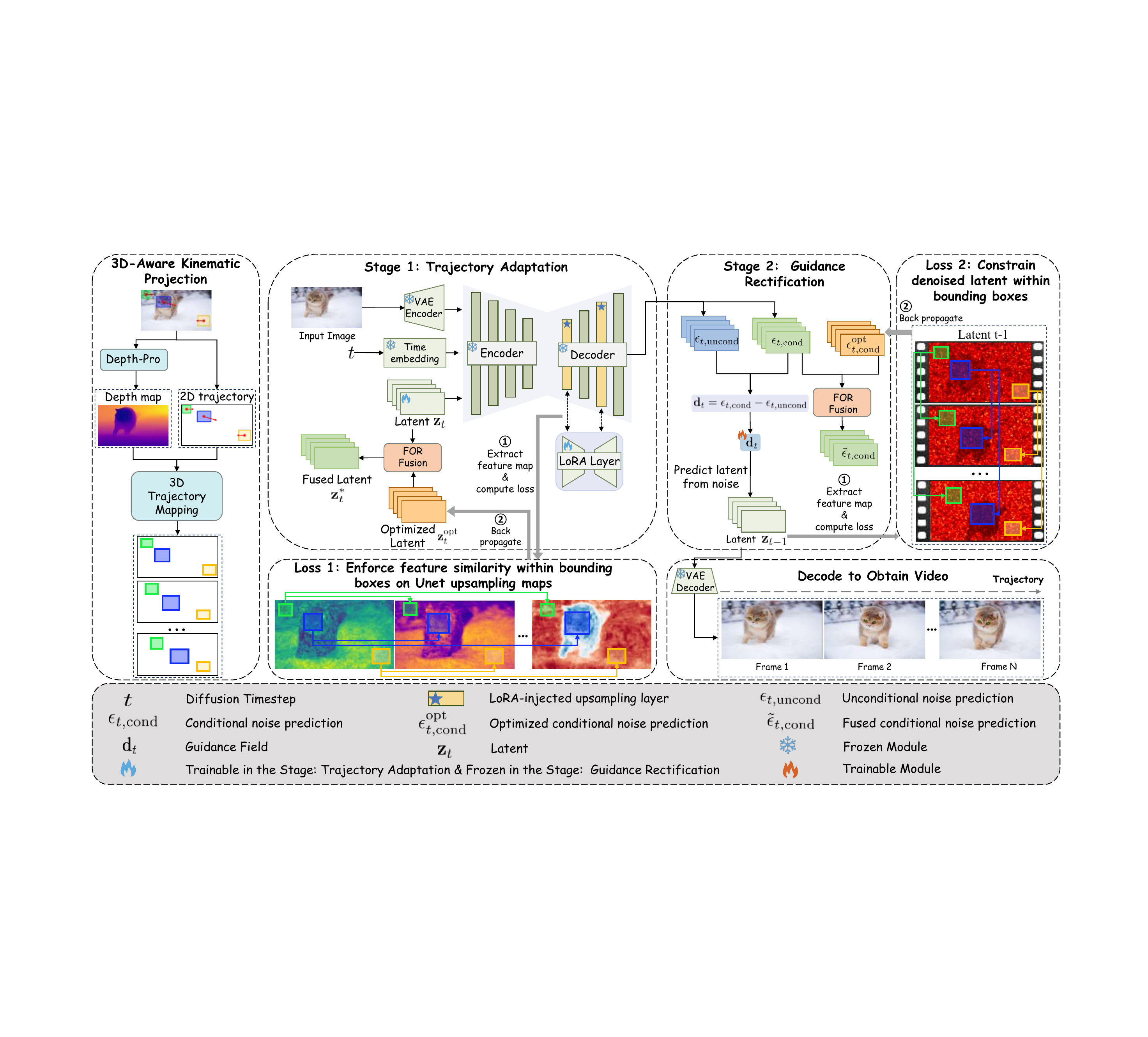} 
\caption{
An overview of our zero-shot trajectory-guided video generation framework.
Our method optimizes a pre-trained video diffusion model at specific denoising timesteps via two key stages. 
First, \textbf{Test-Time Training (TTT)} adapts the latent state and an ephemeral adapter to maintain semantic consistency along the trajectory. 
Second, \textbf{Guidance Field Rectification} refines the denoising direction using a one-step lookahead optimization to ensure precise path execution.
}
\label{pipeline}
\vspace{-1.2em}
\end{figure*}

\section{Methodology}
Zo3T is architected to function within the latent space of a pre-trained image-to-video model, such as Stable Video Diffusion (SVD), without requiring any fine-tuning. As illustrated in Figure~\ref{pipeline}, our method comprises three main components. First, we project user-defined 2D trajectories into a pseudo-3D space, leveraging monocular depth estimation to derive perspective-aware affine transformations. Second, we perform a \textbf{Test-Time Training (TTT)} to co-adapt the latent state and an ephemeral model adapter to enforce feature-space kinematic constraints during the structure-forming stages of denoising.
Finally, the \textbf{Guidance Field Rectification} strategy refines the denoising direction via a one-step lookahead optimization, ensuring precise denoising path execution for trajectory control.

\subsection{3D-Aware Kinematic Projection}
\label{3D_tra}
To endow the model with a foundational awareness of 3D space, we establish a kinematic model to simulate perspective projection. Given a pre-trained monocular depth estimation network $\mathcal{D}_{\text{depth}}$, we compute the depth map $M_D = \mathcal{D}_{\text{depth}}(I_0)$ of the initial frame $I_0$. 
For an object centered at pixel coordinates $\mathbf{p}_k = (u_k, v_k)$ in frame $k$, we approximate its depth as $d_k = M_D(\mathbf{p}_k)$. 
The perspective scaling factor $\sigma_k$ relative to the initial frame is then derived from a pinhole camera approximation: $\sigma_k = d_0 / d_k$.
This scaling allows us to define a time-varying affine transformation $\mathbf{A}_k$ that maps the initial bounding box $\mathcal{B}_0$ to its perspective-aware counterpart $\mathcal{B}_k$ for each frame $k$:
\begin{equation}
    \mathbf{A}_k = 
    \begin{pmatrix}
        \sigma_k & 0 & u_k - \sigma_k u_0 \\
        0 & \sigma_k & v_k - \sigma_k v_0 \\
        0 & 0 & 1
    \end{pmatrix}.
    \label{eq:affine_transform}
\end{equation}
From these transformations, we derive a set of time-varying binary masks $\mathcal{M}_k(\mathbf{p}) = [\mathbf{p} \in \mathbf{A}_k \mathcal{B}_0]$, which serve as the kinematic prior for subsequent optimization stages.

\subsection{Zero-Shot Tajectory-Guided Adaptation via Test-Time Training}
\label{TTT}
Video diffusion models learn a data distribution $p_\theta(\mathbf{x}_0)$ by reversing a noise-corruption process over $T$ timesteps. 
To reduce computational cost, latent diffusion models like SVD operate in a compressed latent space, where a Variational Autoencoder (VAE) maps a raw video $\mathbf{x}_0$ to a latent representation $\mathbf{z}_0$.
The reverse process is driven by a denoiser $\epsilon_\theta(\mathbf{z}_t, t, c)$ trained to predict the noise at timestep $t$. 
This network, which implicitly approximates the score function $\nabla_{\mathbf{z}_t} \log p_t(\mathbf{z}_t|c)$, defines the vector field that guides samples back towards the learned data manifold.

Given a well-defined kinematic prior $\mathcal{M}_k(\mathbf{p})$ that provides per-frame target regions for guidance, the primary challenge in zero-shot trajectory-guided generation lies in enforcing motion constraints within these regions without disrupting the delicate balance between the latent state $\mathbf{z}_t$ and the denoising network $\epsilon_\theta$. Prior zero-shot methods~\cite{namekata2024sg,zhang2025motionawareconceptalignmentconsistent} often employ ``hard edits'' to the latents $\mathbf{z}_t$ to realize object movement. Such approaches can push the latent state off the learned data manifold into configurations the pre-trained model cannot coherently interpret, resulting in significant generative artifacts.

To resolve this challenge, we introduce a \textbf{Test-Time Training (TTT)} paradigm that performs a ``soft adaptation" of both the data and the model. Inspired by previous drag-based control methods~\cite{namekata2024sg,dav24,pan2023drag}, we leverage the cross-frame similarity of deep features within target regions as trajectory guiding signals.
At a specific denoising step $t\in T$, we introduce an ephemeral low-rank perturbation $\Delta\theta_{\text{LoRA}}$ to the model's parameters and co-adapt it with the latent state $\mathbf{z}_t$ by enforcing feature consistency within the target regions. This TTT process seeks a new state-operator equilibrium $(\mathbf{z}_t^*, \theta'=\theta\oplus\Delta\theta_{\text{LoRA}})$ while minimizing a test-time objective functional, $\mathcal{J}_{\text{TTT}}$:
\begin{equation}
    (\mathbf{z}_t^*, \theta_{\text{LoRA}}^*) = \text{argmin}_{\mathbf{z}_t, \theta_{\text{LoRA}}} \mathcal{J}_{\text{TTT}}(\mathbf{z}_t, \theta').
    \label{eq:ttt_objective}
\end{equation}

The test-time objective $\mathcal{J}_{\text{TTT}}(\mathbf{z}_t, \theta')$ is a feature-space consistency loss that enforces alignment between the features of a tracked object in subsequent frames and its features in the first frame:
\begin{equation}
\label{eq:ttt_loss_aligned}
\begin{aligned}
    \mathcal{J}_{\text{TTT}} = \sum_{b=1}^{M} \sum_{l \in \mathcal{L}} w_l \sum_{k=2}^{N_f} \Biggl\| \mathcal{G}_{b} \odot \Biggl( & F_{l,k}[\mathcal{M}_{b,k}] \\
     - & \left(F_{l,1}\right)_{\text{frozen}}[\mathcal{M}_{b,1}] \Biggr) \Biggr\|_F^2.
\end{aligned}
\end{equation}
Here, $F_{l,k} = \phi_{l}(\mathbf{z}_t; \theta')_k$ represents the full feature map from layer $l$ at frame $k$, produced by the adapted model $\theta'$. The notation $[\mathcal{M}_{b,k}]$ denotes cropping the feature map to to the region defined by the mask for object $b$ at frame $k$. Since the 3D-aware kinematic projection can yields regions of varying sizes across frames, we interpolate all cropped feature maps to a uniform size before computing the pixel-wise loss. The reference features from the first frame, $(F_{l,1})_{\text{frozen}}$, are detached from the computation graph (stop-gradient) to provide a stable optimization target. To mitigate boundary effects, a Gaussian heatmap $\mathcal{G}_{b}$ focuses the loss on the object's center via the Hadamard product ($\odot$)~\cite{namekata2024sg}. For the selection of feature layers $\mathcal{L}$, we follow the methodology of~\cite{namekata2024sg,wu2023tune}, which has been proven to yield superior cross-frame spatial-semantic alignment. The TTT process, implemented as a short iterative loop at specific denoising steps, ensures the resulting trajectory is both geometrically accurate and semantically coherent. 
Trajectory-Guided TTT steers the model, via a semantic loss, to adjust the content layout in subsequent frames, thereby realizing controllable object motion.

\vspace{-1em}
\subsection{Guidance Field Rectification via One-Step Lookahead}
\label{GFR}

While the TTT process successfully constrains the solution space to ensure semantic consistency along the trajectory, it does not prescribe an optimal path within that space. The resulting denoising direction is therefore not guaranteed to be the most efficient or precise for trajectory adherence. Furthermore, the strict spatial consistency enforced by TTT can sometimes lead to unnatural or rigid visual results. To resolve this, we introduce a lookahead-based guidance rectification strategy inspired by the principles of Classifier-Free Guidance (CFG)~\cite{ho2022cfg}.

Within the CFG framework, the denoising direction is synthesized from a conditional prediction, which drives the generation towards the control target, and an unconditional prediction, which maintains generative naturality. Their difference can be interpreted as the \textbf{guidance field}, $\mathbf{d}_t = \bm{\epsilon}_t^{\text{cond}} - \bm{\epsilon}_t^{\text{uncond}}$, encapsulating the directional force for control~\cite{li2025flowdirector}. However, this standard guidance field provides a semantically plausible path of least resistance, which may not align with the globally optimal path for precise trajectory control.

To overcome this challenge, we propose Guidance Field Rectification (GFR) by optimizing it with foresight. This is achieved via a \textbf{one-step lookahead} optimization, where our objective is to find an optimal field, $\mathbf{d}_t^*$, that steers the probability flow ODE in a direction that is demonstrably optimal for the subsequent state's alignment with the kinematic prior. This lookahead rectification stage contrasts fundamentally with TTT: whereas TTT adapts the state-operator pair $(\mathbf{z}_t, \theta')$ in \textit{feature space} to ensure representational capacity, this stage rectifies the guidance field $\mathbf{d}_t$ by minimizing a kinematic loss in \textit{latent space}. To this end, we define a cost functional, $\mathcal{J}_{\text{guide}}$, that measures the kinematic inconsistency of the \textit{prospective} latent state at timestep $t-1$:
\begin{small}
\begin{equation}
\label{eq:guide_functional_aligned}
\begin{aligned}
    \mathcal{J}_{\text{guide}}(\mathbf{d}) = \frac{1}{2} \sum_{b=1}^{M} \sum_{k=2}^{N_f} &\int_{\mathcal{B}_{b,k}} \mathcal{G}_{b}(\mathbf{p}) \Biggl\| \text{Pooling}\bigl(\mathcal{P}(\mathbf{z}_t^*, \mathbf{d})_k[\mathcal{B}_{b,1}]\bigr) \\
    & - \text{Pooling}\bigl(\mathcal{P}(\mathbf{z}_t^*, \mathbf{d})_1[\mathcal{B}_{b,1}]\bigr)_{\text{frozen}} \Biggr\|_2^2 \, d\mathbf{p}.
\end{aligned}
\end{equation}
\end{small}
Here, $\mathcal{P}(\mathbf{z}_t^*, \mathbf{d})$ denotes the one-step DDIM solver that evolves the current state $\mathbf{z}_t^*$ to a prospective state $\mathbf{z}_{t-1}$ using a candidate field $\mathbf{d}$. Unlike the pixel-wise MSE loss used in our TTT stage, this guidance loss is computed on spatially pooled features, capturing a holistic representation of the target region. This form of supervision serves a dual purpose: It further guides the object towards the target trajectory while relaxing the strict spatial constraints of TTT, thereby mitigating potential unnatural or rigid visual generation. The optimal field $\mathbf{d}_t^*$ is found by evolving an initial field, $\mathbf{d}_0 = \mathbf{d}_t$, along the negative gradient flow of this functional, as described by $\partial \mathbf{d}/\partial \tau = - \nabla_{\mathbf{d}} \mathcal{J}_{\text{guide}}(\mathbf{d})$.
The final, rectified conditional noise estimate is then recomposed:
\begin{equation}
    \bm{\epsilon}_t^{\text{cond, opt}} = \bm{\epsilon}_t^{\text{uncond}} + \mathbf{d}_t^*.
\end{equation}
This lookahead rectification strategy enables precise steering of the generative trajectory by maximizing its alignment with motion constraints, achieved with minimal overhead via a short, localized optimization loop.

\subsection{Inference Strategies}
\noindent \textbf{Fidelity Preservation via Fourier Orthogonal Recomposition.}
While our iterative optimizations achieve effective trajectory control, they risk causing a distributional shift from the diffusion model's learned prior, leading perceptual quality degradation. Recognizing that motion is predominantly encoded in low-frequency components~\cite{FreeInit}, we introduce the Fourier Orthogonal Recomposition (FOR) strategy. This method preserves the high-frequency textural details from the original signal $\mathbf{x}$, which are crucial for maintaining distributional fidelity, while integrating the low-frequency structural modifications from our optimization $\mathbf{x}^*$. We define a generalized fusion operator $\mathcal{M}_{fuse}$ as:
\begin{equation}
\label{eq:orthogonal_recomposition}
    \mathcal{M}_{fuse}(\mathbf{x}^*, \mathbf{x}) \triangleq \mathcal{F}^{-1} \bigl( \mathcal{F}(\mathbf{x}^*) \odot \mathbf{H}_\gamma + \mathcal{F}(\mathbf{x}) \odot (1 - \mathbf{H}_\gamma) \bigr),
\end{equation}
where $\mathcal{F}$ is the Fourier Transform and $\mathbf{H}_\gamma$ is an ideal low-pass filter mask.
This operator is applied twice at the  denoising step $t$: first to the latent states ($\tilde{\mathbf{z}}_t = \mathcal{M}_{fuse}(\mathbf{z}_t^*, \mathbf{z}_t)$) following TTT, and second to conditional noise predictions ($\tilde{\bm{\epsilon}}_{t, \text{cond}} = \mathcal{M}_{fuse}(\bm{\epsilon}_{t, \text{cond}}^*, \bm{\epsilon}_{t, \text{cond}})$) after guidance rectification. 
By drawing low-frequency structure from optimization and high-frequency details from the original signal, the module preserves control accuracy and visual fidelity.

\noindent \textbf{Selective Timestep Optimization.}
The coarse structure and motion dynamics of the generated video are largely determined during the early stages of the denoising process~\cite{DBLP:journals/corr/abs-2303-02490}. To maximize computational efficiency and ensure structural stability, we apply our trajectory guidance optimizations selectively. Specifically, both the TTT adaptation and GFR are performed only during the early-to-mid denoising phase, from timestep $t=45$ down to $t=30$ in a 50-step schedule. This selective application aligns with established practices in the controllable generation literature~\cite{namekata2024sg}, striking an effective balance between precise control and generative quality.

\begin{figure*}[t]
\centering
\setlength{\abovecaptionskip}{-0.03em}   
\includegraphics[width=\textwidth]{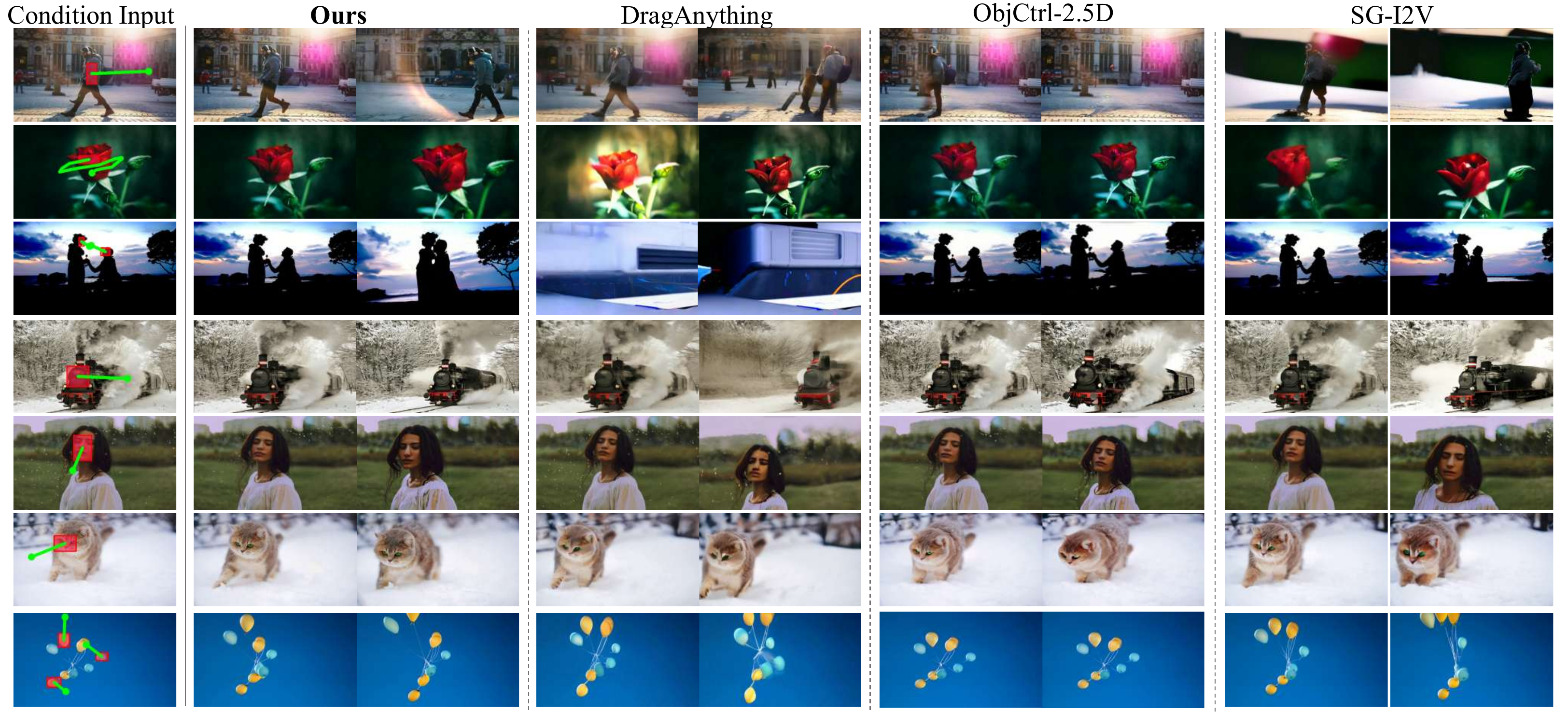} 
\caption{Qualitative Comparison with SOTA Methods.} 
\label{fig:qualitative}
\vspace{-1em}
\end{figure*}

\begin{table*}[t]
\centering
\resizebox{\textwidth}{!}{
\begin{tabular}{
    l 
    c 
    S[table-format=3.2] 
    S[table-format=3.2] 
    S[table-format=2.2] 
    S[table-format=1.4] 
    S[table-format=1.4] 
    S[table-format=1.4] 
    S[table-format=1.4] 
    c 
    l 
}
\toprule
\textbf{Method} & 
\multicolumn{1}{c}{\textbf{Zero-shot}} & 
\multicolumn{1}{c}{\textbf{FID} ($\downarrow$)} & 
\multicolumn{1}{c}{\textbf{FVD} ($\downarrow$)} & 
\multicolumn{1}{c}{\textbf{ObjMC} ($\downarrow$)} & 
\multicolumn{1}{c}{\begin{tabular}[c]{@{}c@{}}\textbf{Subject}\\\textbf{Consist.}($\uparrow$)\end{tabular}} & 
\multicolumn{1}{c}{\begin{tabular}[c]{@{}c@{}}\textbf{Bkg.}\\\textbf{Consist.}($\uparrow$)\end{tabular}} & 
\multicolumn{1}{c}{\begin{tabular}[c]{@{}c@{}}\textbf{Aesthetic}\\\textbf{Quality}($\uparrow$)\end{tabular}} & 
\multicolumn{1}{c}{\begin{tabular}[c]{@{}c@{}}\textbf{Imaging}\\\textbf{Quality}($\uparrow$)\end{tabular}} & 
\multicolumn{1}{c}{\textbf{Resolution}} & 
\textbf{Backbone} \\ 
\midrule

DragNUWA~\cite{yin2023dragnuwa} & {\ding{55}}  & 126.31 & 251.04 & \textbf{10.84} & 0.9177 & 0.9272  & 0.5412 & 0.5433 & 320 $\times$ 576 & SVD \\
DragAnything~\cite{wu2024draganything} & {\ding{55}} & 119.07 & 266.42 & 11.64 & 0.9204 & 0.9262 & 0.5469 & 0.5711 & 320 $\times$ 576 & SVD \\
LeviTor~\cite{levitor} & {\ding{55}} & 79.41 & 207.23 & 12.06 & 0.9482 & 0.9511 & 0.6417 & 0.6314 & 288 $\times$ 512 & SVD \\
\midrule

FreeTraj~\cite{qiu2024freetraj} & {$\checkmark$} & 92.12 & 230.21 & 31.63 & 0.9236 & 0.9281 & 0.5841 & 0.5903 & 320 $\times$ 512 & VideoCrafter2 \\
ObjCtrl-2.5D~\cite{objctrl2.5d} & {$\checkmark$} & 81.06 & 212.17 & 18.72 & 0.9512 & 0.9544 & 0.6344 & 0.6429 & 320 $\times$ 576 & SVD \\
SG-I2V~\cite{namekata2024sg} & {$\checkmark$} &  79.36 & 209.53 & 14.43 & 0.9448 & 0.9517 & 0.6370 & 0.6317  & 576 $\times$ 1024 &  SVD\\
\midrule

\textbf{Ours} & {$\checkmark$} & \textbf{74.83} &\textbf{197.63} & 12.74 & \textbf{0.9760} & \textbf{0.9682} & \textbf{0.6779} & \textbf{0.6820}  & 576 $\times$ 1024 & SVD \\
\bottomrule
\end{tabular}
} 
\vspace{-1em}
\caption{
    Quantitative comparison on the VIPSeg dataset.
    Despite being a zero-shot approach, our method achieves a small gap in motion fidelity (ObjMC) compared to supervised baselines, without degrading video quality (FID, FVD). 
    Furthermore, our approach outperforms other zero-shot baselines across all metrics. 
    ($\downarrow$) indicates lower is better, ($\uparrow$) indicates higher is better.
}
\label{tab:quantitative_comparison}
\vspace{-1em}
\end{table*}

\section{Experiments}
\subsection{Experiment Settings}
\noindent \textbf{Implementation Details.} Our framework is built upon the Stable Video Diffusion (SVD)~\cite{blattmann2023stable}, initialized with its official pre-trained weights. All experiments are configured to generate 14-frame videos at a resolution of 576$\times$1024. The optimization process employs the Adam optimizer with learning rates of 0.25, 0.01, and 0.05 for the latent, LoRA weights, and the guidance field, respectively. We employ DepthPro~\cite{depthpro} for monocular depth estimation, the output of which is used to construct 3D-aware trajectories and the corresponding affine transformations of target regions. For the Fourier Orthogonal Recomposition (FOR) module, the low-pass filter's cutoff frequency is set to 0.6. A comprehensive ablation analysis of these experimental parameters is detailed in the \textbf{\textit{Appendix}}.

\noindent \textbf{Evaluation Metrics.}
We conduct a comprehensive evaluation using both quantitative metrics and a user study. For quantitative assessment, we measure \textbf{\textit{Video Quality}} via FID, FVD, and four key VBench~\cite{huang2023vbench} metrics (subject consistency, background consistency, aesthetic quality, and imaging quality), and \textbf{\textit{Motion Control Accuracy}} using Object Motion Conformity (ObjMC)~\cite{namekata2024sg}—the Euclidean distance between target and actual trajectories. In parallel, our user study involved six expert evaluators who assessed 50 videos generated from a custom dataset of images, each with a predefined trajectory.

\noindent \textbf{Evaluation Datasets.} For a fair comparison, we adopt the open-source dataset used in~\cite{wu2024draganything}. Recognizing that many baseline I2V models excel on common, in-distribution trajectories~\cite{objctrl2.5d}, we augment this set by applying a ``mirroring" transformation to most paths, resulting in a more challenging set of 143 samples featuring more diverse and less common motion paths. Following the protocols in~\cite{wu2024draganything}, FID and FVD metrics are computed on the VIPSeg~\cite{vipseg} dataset.

\subsection{Versatile Trajectory Control Modes}
Zo3T provides flexible trajectory control for any designated entity via a bounding box. This includes enforcing regional stillness by setting the trajectory vector to zero. Figure~\ref{fig:teaser} showcases two primary control modalities offered by our framework. 
(1) \textbf{Object Motion Trajectory Control:}
By defining a bounding box around a target object and specifying a desired trajectory, Zo3T can direct the object's movement along the prescribed path. As shown in Figure~\ref{fig:teaser}(a), Zo3T can effectively control both foreground (a1) and background (a2) objects, ensuring they accurately track their respective trajectories. 
(2) \textbf{Camera Motion Trajectory Control:}
Camera motion is simulated by defining a bounding box over a background region and assigning it a trajectory inverse to the intended camera movement. As shown in Figure~\ref{fig:teaser}(b), Zo3T can can effectively perform complex camera operations, including dolly zooms (b1) and pans (b2).

\begin{figure*}[t]
\centering
\setlength{\abovecaptionskip}{-0.03em}   
\includegraphics[width=\textwidth]{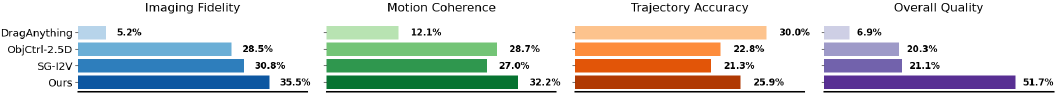} 
\caption{User Study. The majority of participants preferred
the results obtained by our method over both training-free and
training-based methods, attributing this preference to its better trajectory alignment and more natural motion generation }
\label{fig:user_study}
\vspace{-1.5em}
\end{figure*}

\begin{figure}[t]
\centering
\setlength{\abovecaptionskip}{-0.03em}   
\includegraphics[width=0.96\columnwidth]{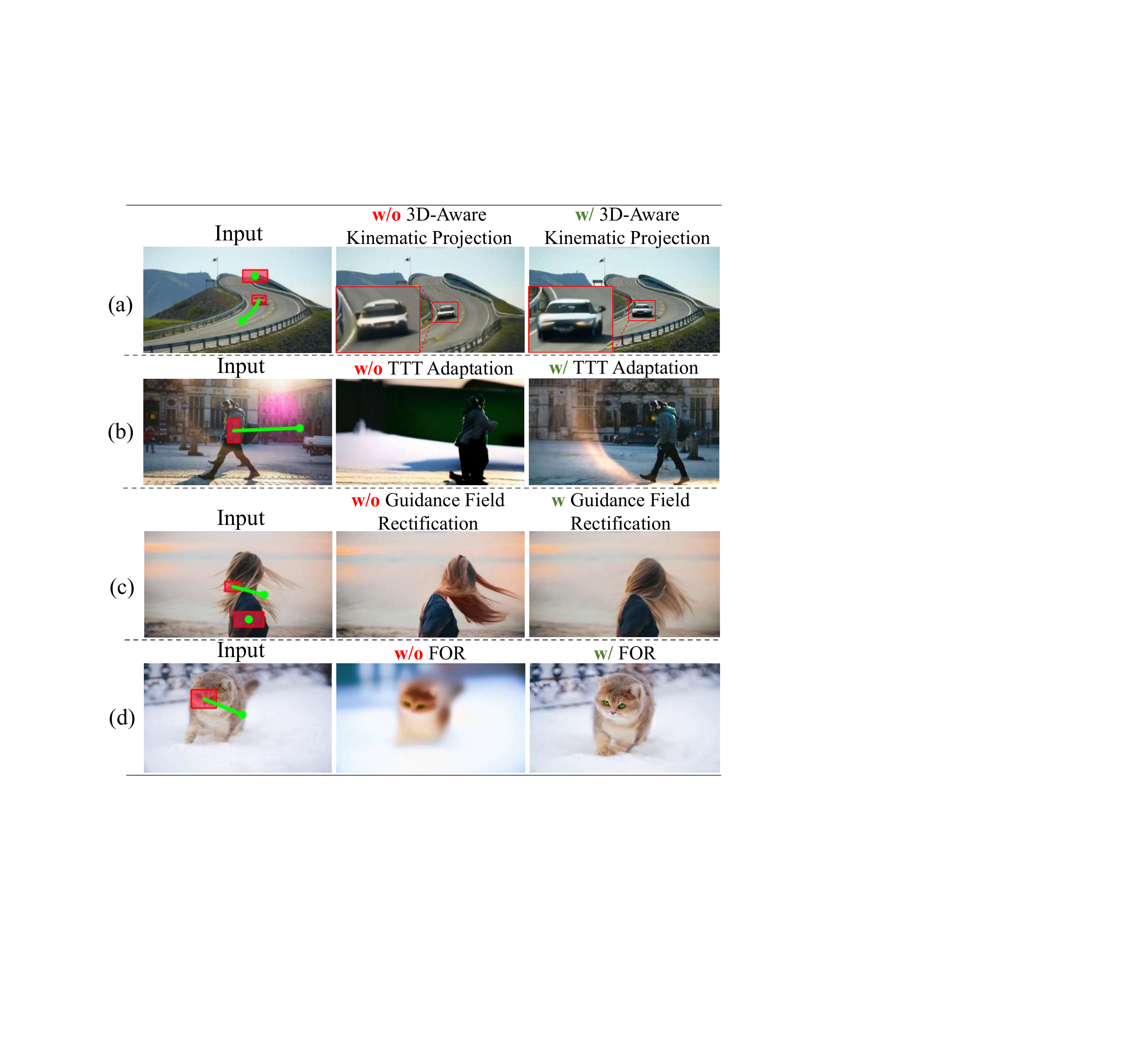} 
\caption{Qualitative Ablation Study}
\label{fig:ablation_qualitative}
\vspace{-1.7em}
\end{figure}

\vspace{-0.5em}
\subsection{Comparisons with State-of-the-Art Methods}
We perform a comprehensive evaluation by comparing Zo3T with SOTA supervised and zero-shot baselines. For supervised baselines, we select DragNUWA~\cite{yin2023dragnuwa}, DragAnything~\cite{wu2024draganything} and LeviTor~\cite{levitor}. The zero-shot methods include SG-I2V~\cite{namekata2024sg}, ObjCtrl-2.5D~\cite{objctrl2.5d}, and an adapted version of FreeTraj~\cite{qiu2024freetraj} for our I2V task, following the procedure in~\cite{namekata2024sg}. For a fair evaluation, all video outputs are resized to a 320×576 resolution to align with the supervised models.

\noindent \textbf{Comparison with Supervised Methods.}
Supervised methods, trained on tracker-derived trajectories, exhibit slightly higher motion precision (ObjMC) but suffer from significant visual artifacts like object distortion, flickering, and style collapse due to their prioritization of trajectory adherence over generative fidelity (Table~\ref{tab:quantitative_comparison}, Figure~\ref{fig:qualitative}). In contrast, our method integrates a Test-Time Training (TTT) generative guidance paradigm with the Fourier Orthogonal Recomposition (FOR) module to restore high-frequency details, yielding superior generation quality (FID, FVD) and more natural motion. Crucially, supervised methods are often limited to lower resolutions (320×576) by the high cost of fine-tuning, whereas our zero-shot framework operates directly at the native high resolution of SVD (576×1024) without requiring any external training or external knowledge.

\begin{table}[ht]
\centering
\resizebox{0.85\columnwidth}{!}{
\begin{tabular}{
    l S[table-format=2.2] S[table-format=3.2] S[table-format=2.2]
}
\toprule
\textbf{Variant} & 
\multicolumn{1}{c}{\textbf{FID} ($\downarrow$)} & 
\multicolumn{1}{c}{\textbf{FVD} ($\downarrow$)} & 
\multicolumn{1}{c}{\textbf{ObjMC} ($\downarrow$)} \\
\midrule
\textbf{Ours (Full Model)} & 
\textbf{74.83} & \textbf{197.63} & \textbf{12.74} \\
\midrule
(a) w/o 3D Projection & 
76.12 & 201.55 & 12.98 \\
(b) w/o TTT Adaptation & 
89.36 & 219.53 & 14.24 \\
(c) w/o GFR & 
78.91 & 205.18 & 13.92 \\
(d) w/o FOR & 
95.45 & 221.09 & 12.81 \\
\bottomrule
\end{tabular}
} 
\vspace{-0.7em}
\caption{
    Ablation study on the core components of our framework.
    We start with our full model and progressively remove each key component. 
    The results demonstrate that every module contributes positively to either video quality (FID/FVD) or motion accuracy (ObjMC).
}
\label{tab:ablation_study}
\vspace{-0.8em}
\end{table}

\noindent \textbf{Comparison with Zero-shot Methods.}
Our method demonstrates clear superiority over all zero-shot baselines.
FreeTraj's reliance on a hand-crafted noise prior degrades fine details and limits control to a coarse-grained level.
ObjCtrl-2.5D, by equating object motion with camera movement, is restricted to single-object control and proves ineffective in complex scenarios requiring coordinated object-background motion. In contrast, our method enables flexible and diverse multi-region trajectory guidance, yielding more accurate and natural motion without additional dependencies.
While SG-I2V strikes a reasonable balance between quality and accuracy, its ``hard editing" in latent space risks manifold deviation, which can lead to artifacts and unnatural results (Figure~\ref{fig:qualitative}). Conversely, our TTT paradigm maintains manifold integrity during trajectory guidance by co-adapting the latent state with an ephemeral LoRA adapter, which is further enhanced by the GFR module to refines the control generation path. This combination enables higher precision while preserving generative fidelity.

\noindent \textbf{User Study.}
We evaluate subjective perceptual quality via a user study where six trained participants performe preference voting on our method against three top competitors based on imaging fidelity, motion coherence, trajectory accuracy, and overall quality.  As shown in Figure~\ref{fig:user_study}, our method secures a clear majority preference of 51.7\% over both SOTA supervised and zero-shot models.

\vspace{-0.5em}
\subsection{Ablation Studies}
We validate each core component via quantitative (Table~\ref{tab:ablation_study}) and qualitative (Figure~\ref{fig:ablation_qualitative}) ablations.

\textbf{\textit{Effectiveness of 3D-Aware Kinematic Projection.}}
We evaluate the impact of our 3D-Aware Kinematic Projection by replacing it with raw 2D trajectory guidance. While this leads to only a minor degradation in quantitative metrics (Table~\ref{tab:ablation_study}), the qualitative impact is far more significant (Figure~\ref{fig:ablation_qualitative}(a)). In scenes with noticeable depth changes, enforcing feature consistency with an unscaled 2D bounding box causes severe object distortions as its perceived size changes. This demonstrates the necessity of our 3D-aware projection for generating physically plausible motion and realistic perspective shifts.

\textbf{\textit{Effectiveness of TTT Tajectory-Guided Adaptation.}}
We then isolate the contribution of our TTT paradigm by removing the LoRA adapter, thereby degenerating our method into a ``hard editing" for the latent $z\_t$, similar to SG-I2V. Although this strategy can still follow the trajectory to some extent, it suffers from a clear drop in generation quality (Table~\ref{tab:ablation_study}) and can even cause the scene to collapse in difficult control cases (Figure~\ref{fig:ablation_qualitative}(b)). This confirms that our TTT framework is critical for resolving the latent-denoiser misalignment, thereby preserving the generative manifold to ensure both high-fidelity video and precise control.

\textbf{\textit{Effectiveness of Guidance Field Rectification.}}
Next, we remove the GFR module from our full framework . The quantitative results in Table~\ref{tab:ablation_study} show a degradation across all metrics, indicating that TTT's motion guidance benefits from the refinement offered by GFR. Qualitatively, by using a one-step lookahead with a holistic feature loss, GFR relaxes the strict spatial constraints imposed by TTT's pixel-wise loss. This mitigates unnatural rigidity and promotes more organic motion, demonstrating GFR's crucial role in harmonizing trajectory accuracy with natural dynamics.

 \textbf{\textit{Effectiveness of Fourier Orthogonal Recomposition.}}
Finally, we evaluate the impact of the FOR module by removing it from the full model. As shown in Table~\ref{tab:ablation_study}, this removal has a negligible effect on motion accuracy (ObjMC) but significantly worsens video quality, as reflected by the higher FID and FVD scores. The qualitative comparison in Figure~\ref{fig:ablation_qualitative}(e) confirms this, showing that videos generated without FOR appear overly smooth and lack fine texture details. This validates FOR's essential role in eliminating optimization-induced artifacts and restoring visual fidelity.

\vspace{-0.5em}
\section{Conclusion}
In this paper, we introduce a novel method for trajectory-guided video generation that addresses the critical challenges of spatial ambiguity and generative quality degradation. 
Our core contribution is a two-stage test-time optimization framework. First, we perform test-time training to co-adapt the latent state and a lightweight model adapter, which steers the object along a perspective-aware 3D path while keeping the generation on the learned data manifold. Second, we introduce Guidance Field Rectification, a one-step lookahead optimization that refines the denoising direction for precise path execution. By integrating these optimizations with a Fourier-based fusion strategy to preserve high-frequency details, our method achieves SOTA performance, generating videos with superior motion accuracy and visual fidelity without the need for supervised fine-tuning.

\section*{Acknowledgements}
This work was partly supported by Shenzhen Key Laboratory of next generation interactive media innovative technology (No:ZDSYS20210623092001004).

\bibliography{aaai2026}

\clearpage
\setcounter{page}{1}
\begin{center}
    \textbf{\Huge Supplementary Material}
\end{center}

\begin{figure*}[h]
\centering
\includegraphics[width=\textwidth]{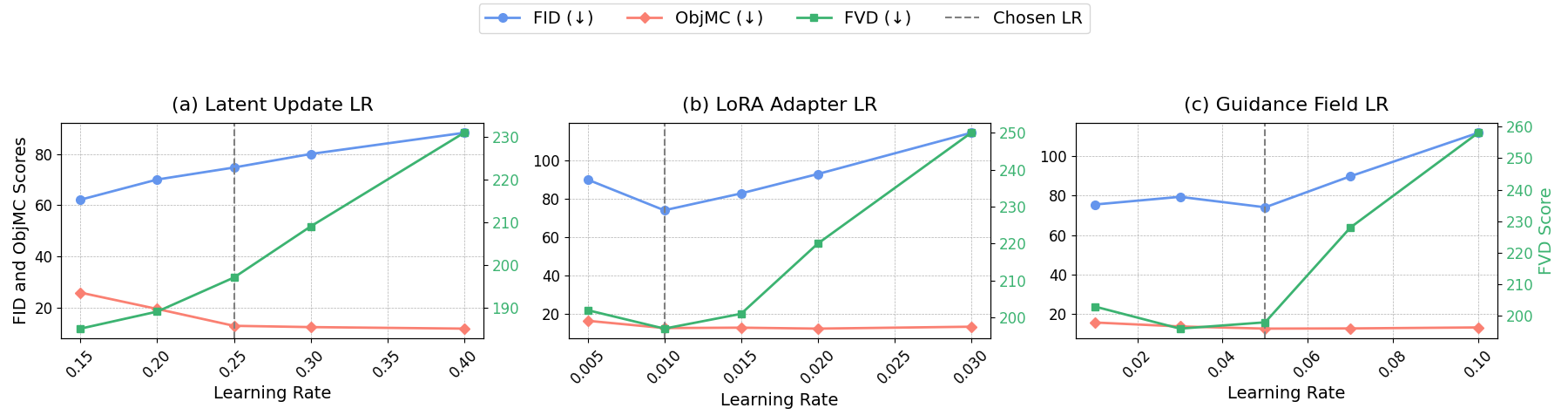} 
\caption{Ablation on optimization learning rates.} 
\label{fig:ablation_lr}
\vspace{-0.5em}
\end{figure*}

\begin{figure*}[h]
\centering
\includegraphics[width=\textwidth]{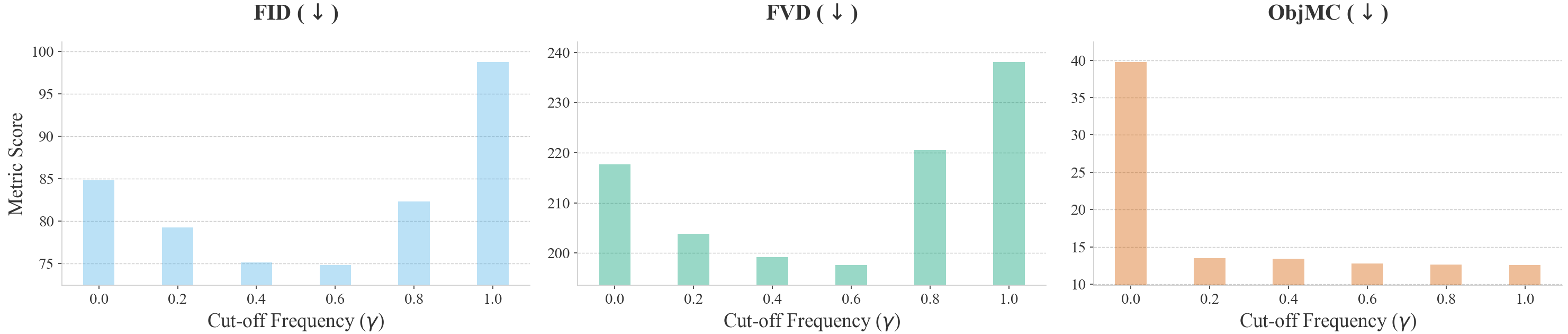} 
\caption{Ablation on Cut-off Frequency in Fourier Orthogonal Recomposition.} 
\label{fig:ablation_fre}
\vspace{-0.5em}
\end{figure*}

\section{Algorithm Overview}
Our proposed method, \ModelName, introduces a zero-shot, two-stage optimization framework that operates within a pre-trained image-to-video diffusion model, as detailed in Algorithm~\ref{alg:main_concise}. The process begins by establishing a \textbf{3D-aware kinematic prior}; we leverage monocular depth estimation on the initial frame to project user-specified 2D trajectories into perspective-correct affine transformations and masks for each frame.

The core of our method unfolds iteratively during a selective range of denoising timesteps (e.g., $t=45$ to $t=30$). At each guided step, we execute a two-stage optimization:
\begin{enumerate}
    \item \textbf{Stage 1: Trajectory-Guided Test-Time Training (TTT).} To address the model-data misalignment caused by direct latent manipulation, we simultaneously optimize both the latent state $\mathbf{z}_t$ and a temporary, lightweight LoRA adapter injected into the U-Net. This co-adaptation is driven by a feature-space consistency loss that enforces the object's appearance to remain constant along its trajectory, effectively aligning the model's internal representations with the edited latent.
    
    \item \textbf{Stage 2: Guidance Field Rectification (GFR).} To ensure precise path execution, we refine the denoising direction itself. We optimize the conditional guidance field $\mathbf{d}_t$ using a one-step lookahead strategy, which seeks a direction that minimizes the kinematic inconsistency of the \textit{prospective} latent state at timestep $t-1$.
\end{enumerate}

To maintain high visual fidelity, the outputs of both optimization stages are integrated back into the main generative path using a \textbf{Fourier Orthogonal Recomposition} strategy. This technique merges the low-frequency structural changes from our guidance with the high-frequency textural details from the original signal. After the guided timesteps, the standard denoising process resumes to generate the final, trajectory-aligned video.

\section{Experiment Setup Details}
\subsection{Evaluation Protocol}
Since existing methods like DragAnything~\cite{wu2024draganything} and DragNUWA are limited to generating low-resolution videos (320$\times$576), we resize all generated videos to this resolution prior to evaluation to ensure a fair comparison. It is worth noting that this normalization protocol does not fully leverage our model's high-resolution synthesis capabilities. Nevertheless, \ModelName still achieves state-of-the-art (SOTA) performance across multiple metrics, which underscores the superiority of our approach. All experiments are conducted on NVIDIA A100 GPUs.

\begin{algorithm}[h]
    \caption{Trajectory-Guided Video Generation via Two-Stage Optimization}
    \label{alg:main_concise}
    \begin{algorithmic}[1]
        \Input Initial frame $I_0$, initial bounding boxes $\{\mathcal{B}_{b,0}\}$, 2D trajectories $\{\mathcal{T}_{2D}^b\}$.
        \Output Trajectory-guided video $\mathcal{V}$.
        
        \State \textbf{Initialize:} Compute kinematic prior $\mathcal{T} = \{\mathbf{A}_{b,k}, \mathcal{M}_{b,k}\}$. $\mathbf{z}_T \sim \mathcal{N}(0, \mathbf{I})$.
        
        \For{$t = T, \dots, 1$}
            \State \textbf{// Stage 1: Trajectory Adaptation (TTT)}
            \State Inject ephemeral LoRA weights $\Delta\theta_{\text{LoRA}}$ into U-Net.
            \State $(\mathbf{z}_t^{\text{opt}}, \Delta\theta_{\text{LoRA}}) \leftarrow$ Iteratively optimize $(\mathbf{z}_t, \Delta\theta_{\text{LoRA}})$ by minimizing $\mathcal{J}_{\text{TTT}}$.
            \State $\mathbf{z}_t^* \leftarrow \mathcal{M}_{fuse}(\mathbf{z}_t^{\text{opt}}, \mathbf{z}_t)$.
            
            \Statex
            \State \textbf{// Stage 2: Guidance Rectification (Lookahead)}
            \State Predict $\bm{\epsilon}_{t, \text{cond}}, \bm{\epsilon}_{t, \text{uncond}}$ from $\mathbf{z}_t^*$ using the adapted U-Net.
            \State $\mathbf{d}_0 \leftarrow \bm{\epsilon}_{t, \text{cond}} - \bm{\epsilon}_{t, \text{uncond}}$.
            \State $\mathbf{d}_t^* \leftarrow$ Iteratively optimize $\mathbf{d}$ from $\mathbf{d}_0$ by minimizing $\mathcal{J}_{\text{guide}}$.
            \State $\bm{\epsilon}_{t, \text{cond}}^{\text{opt}} \leftarrow \bm{\epsilon}_{t, \text{uncond}} + \mathbf{d}_t^*$.
            \State $\tilde{\bm{\epsilon}}_{t, \text{cond}} \leftarrow \mathcal{M}_{fuse}(\bm{\epsilon}_{t, \text{cond}}^{\text{opt}}, \bm{\epsilon}_{t, \text{cond}})$.
            
            \Statex
            \State \textbf{// Denoising Step}
            \State Synthesize final noise $\bm{\epsilon}_t^{\text{final}}$ with rectified conditional estimate $\tilde{\bm{\epsilon}}_{t, \text{cond}}$.
            \State $\mathbf{z}_{t-1} \leftarrow \mathcal{P}(\mathbf{z}_t^*, \bm{\epsilon}_t^{\text{final}})$.
        \EndFor
        
        \State \textbf{Finalize:} $\mathcal{V} \leftarrow \mathcal{D}(\{\mathbf{z}_{k}\}_{k=0}^{N_f-1})$.
    \end{algorithmic}
\end{algorithm}

\subsection{Metric Details}
\paragraph{Fréchet Video Distance (FVD).}
For FVD calculation, we follow the protocol established in prior works~\cite{wu2024draganything,namekata2024sg}. All videos, including our generated results and the ground-truth videos from the VIPSeg dataset~\cite{vipseg}, are resized to 256$\times$256. We employ the ViCLIP~\cite{internvid} model as the feature extractor. The FVD is computed as the Fréchet distance between the distributions of real and generated video features:
\begin{equation}
\label{eq:fvd}
\text{FVD} = \|\mu_r - \mu_g\|_2^2 + \text{Tr}(\Sigma_r + \Sigma_g - 2(\Sigma_r\Sigma_g)^{1/2}),
\end{equation}
where $(\mu_r, \Sigma_r)$ and $(\mu_g, \Sigma_g)$ represent the mean and covariance of feature embeddings from the real and generated videos, respectively.

\paragraph{Object Motion Consistency (ObjMC).}
Following the approach in DragNUWA~\cite{yin2023dragnuwa}, we evaluate motion fidelity using the Object Motion Consistency (ObjMC) metric. This metric is defined as the average per-frame Euclidean distance between the tracked trajectory and the ground-truth trajectory. The trajectory in the generated video is estimated using Co-Tracker~\cite{cotracker}.

It is crucial to note that a high ObjMC score does not necessarily correlate with precise trajectory control or high visual quality. We observe that in videos generated by supervised methods like DragAnything and DragNUWA, the tracked object frequently suffers from severe distortion or complete structural collapse during motion. Despite the object losing its coherent features and shape, Co-Tracker can often continue to track it accurately. This results in a paradoxical scenario where a video of poor visual quality can still achieve a deceptively high ObjMC score.

\section{Ablation Experiments on Parameter Selection}
\subsection{Learning Rates}
Figure~\ref{fig:ablation_lr} illustrates the impact of learning rates for the latent updates, LoRA adapter optimization, and guidance field rectification on our three key metrics. We observe a clear trade-off: higher learning rates tend to degrade video quality, resulting in increased FID and FVD scores, whereas lower learning rates reduce motion control accuracy, leading to a higher ObjMC. The figure marks our chosen learning rates, which strike the best balance between generative fidelity and trajectory adherence.

\subsection{Cut-off Frequency in Fourier Orthogonal Recomposition}
In Figure~\ref{fig:ablation_fre}, we analyze the effect of the low-pass filter's cut-off frequency ($\gamma$) in our Fourier Orthogonal Recomposition. 
Simply retaining the fully optimized latent variable ($\gamma=1$) results in a noticeable degradation of visual quality, as evidenced by higher FID and FVD metrics. 
However, discarding some high-frequency components has a minimal effect on motion control while eliminating most artifacts. 
We therefore select $\gamma=0.6$ as the optimal balance point to ensure video fidelity.

\subsection{Inference Cost}
Our method generates 14-frame videos at a 576$\times$1024 resolution using 50 DDPM sampling steps. The optimization process is selectively applied from timestep $t=45$ down to $t=30$. On a single NVIDIA A100 (80GB) GPU, the average inference time is approximately 175 seconds, which varies based on the number of trajectory conditions. The backpropagation required for this optimization leads to a peak memory footprint of roughly 50 GB. We note that these computational costs could be substantially mitigated by integrating optimized attention mechanisms, such as FlashAttention~\cite{flashattention} or SagAttention~\cite{sageattention}.

\section{Additional Visual Results}
We present additional qualitative results across several challenging trajectory-guided scenarios. Across all examples, \ModelName demonstrates a superior balance between precise trajectory adherence, 3D-aware physical realism, and high-fidelity visual generation. In contrast, supervised methods like DragAnything often struggle to maintain object integrity, leading to noticeable distortion and structural collapse as the object moves. Meanwhile, other training-free methods that directly manipulate latents may fail to follow the intended trajectory due to insufficient control. This approach also frequently introduces visual artifacts due to off-manifold optimization, degrading the object's identity and texture. These results underscore \ModelName's unique capability to execute complex motion commands with high accuracy while preserving the visual and structural integrity of the scene, a testament to our dual-stage optimization and 3D-aware kinematic modeling.

\begin{figure*}[t]
    \centering
    \setlength{\abovecaptionskip}{2mm}

    \begin{subfigure}{\textwidth}
        \centering
        \includegraphics[width=0.98\textwidth]{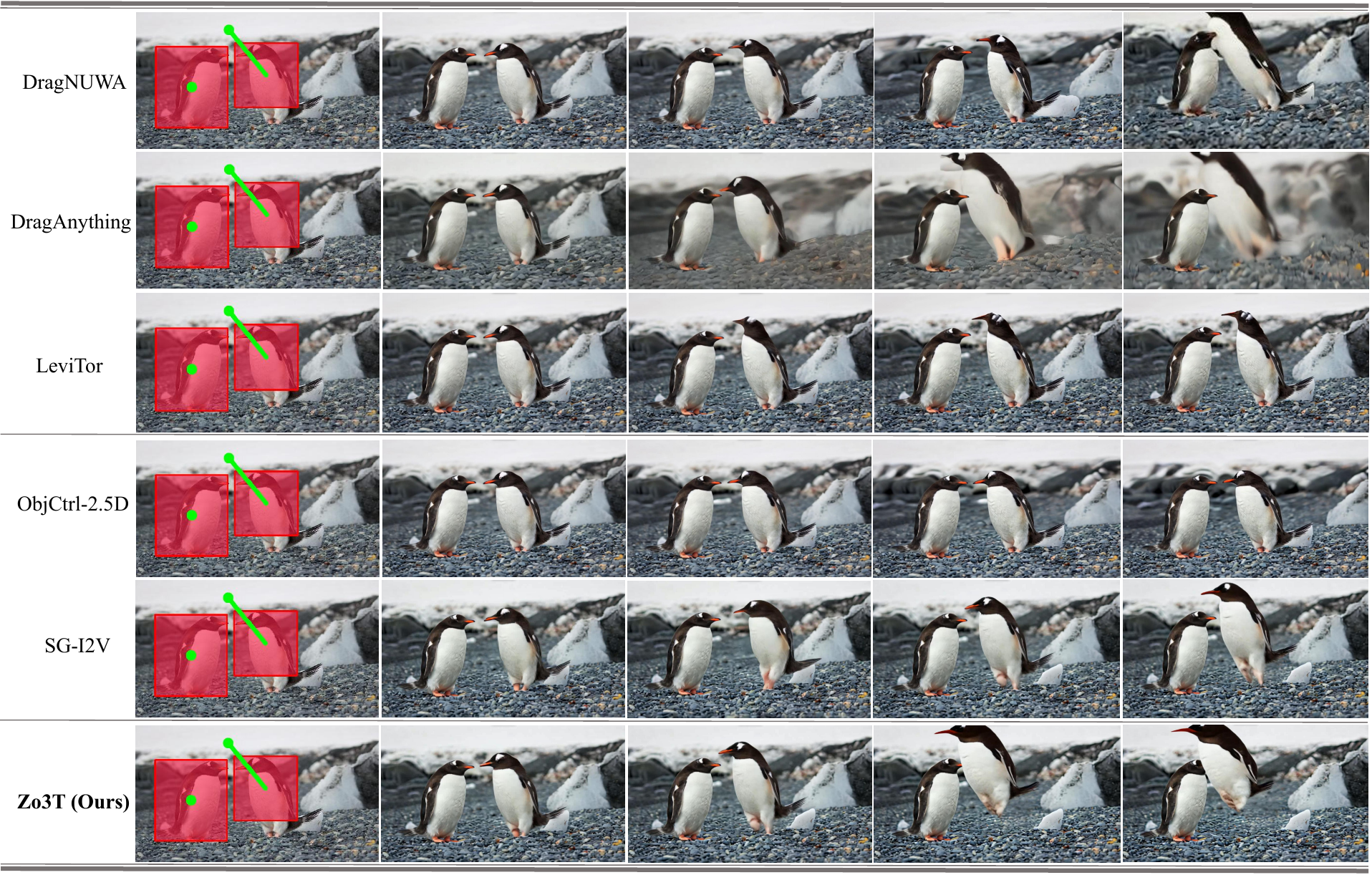}
        \label{fig:vis_sub1}
    \end{subfigure}
    
    \begin{subfigure}{\textwidth}
        \centering
        \includegraphics[width=0.98\textwidth]{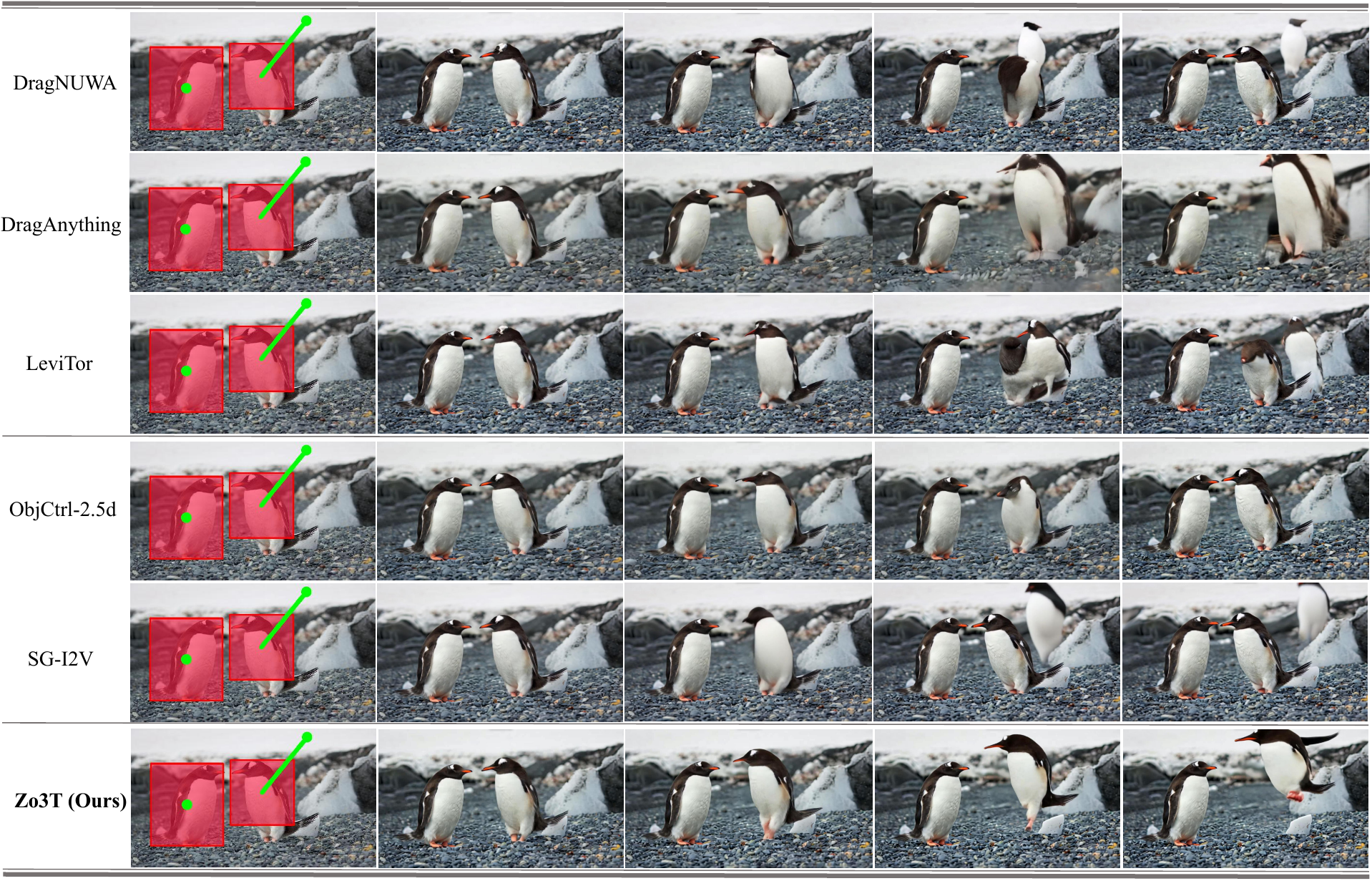}
        \label{fig:vis_sub2}
    \end{subfigure}

    \caption{More Qualitative Comparisons. Our method demonstrates robust performance in various scenarios.  Best viewed zoomed-in.}
    \label{fig:vis_res_combined}
    \vspace{-1em}
\end{figure*}

\begin{figure*}[t]
\centering
\includegraphics[width=0.95\textwidth]{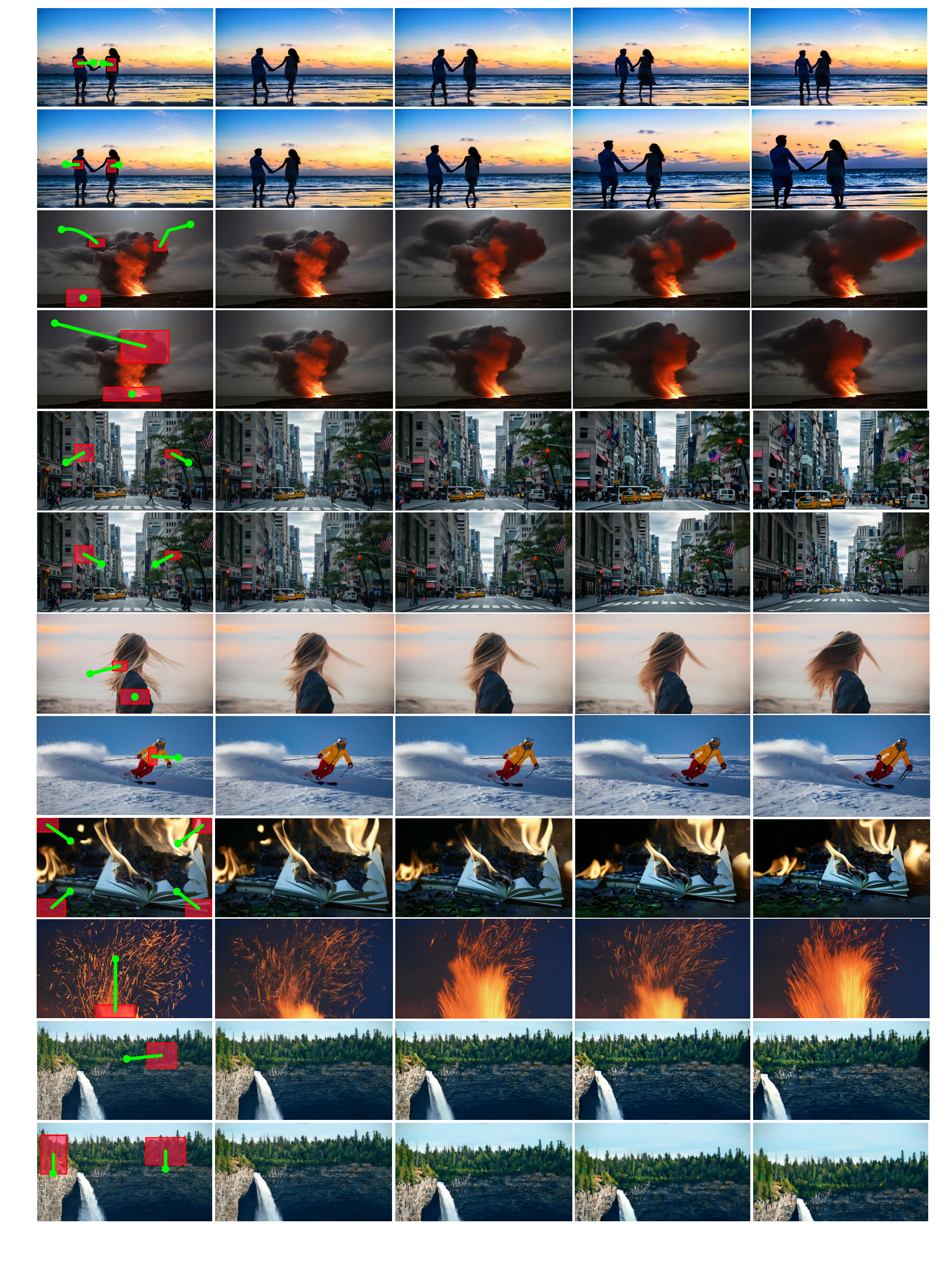} 
\caption{More Qualitative Results. Our method performs precise and semantically faithful edits
 while preserving the spatial content and motion dynamics of unedited regions. These results exhibit strong alignment with the trajectory instructions, high visual fidelity, and consistent temporal coherence across frames. Best viewed zoomed-in.} 
\label{fig:vis_res2}
\vspace{-1em}
\end{figure*}

\end{document}